\documentclass[10pt,twocolumn]{wlscirep}
\usepackage[utf8]{inputenc} % allow utf-8 input
\usepackage[T1]{fontenc}    % use 8-bit T1 fonts
\usepackage{lettrine}
\usepackage{multicol}

\usepackage{caption}
\usepackage{graphicx}
\usepackage{float} 
\usepackage{subfigure}
\usepackage{subcaption}

\usepackage{hyperref}       % hyperlinks
\usepackage{url}            % simple URL typesetting
\usepackage{booktabs}
\usepackage[normalem]{ulem}
\useunder{\uline}{\ul}{}
\usepackage{amsfonts}       % blackboard math symbols
\usepackage{nicefrac}       % compact symbols for 1/2, etc.
\usepackage{microtype}      % microtypography
\usepackage{xcolor}         % colors
\usepackage{amsmath}
\usepackage{cases}
\usepackage[ruled,linesnumbered]{algorithm2e}

\title {Combining Knowledge Graph and LLMs for Enhanced Zero-shot Visual Question Answering}

\author[1,*]{Qian Tao}
\author[1]{Xiaoyang Fan}
\author[1]{Yong Xu}
\author[2]{Xingquan Zhu}
\author[2]{Yufei Tang}
\affil[1]{South China University of Technology, Guangzhou, China}
\affil[2]{Florida Atlantic University, Boca Raton, Florida, USA}
\affil[*]{Corresponding author: taoqian@scut.edu.cn}
\begin{abstract}
Zero-shot visual question answering (ZS-VQA), an emerged critical research area, intends to answer visual questions without providing training samples. Existing research in ZS-VQA has proposed to leverage knowledge graphs or large language models (LLMs), respectively, as external information sources to help VQA model comprehend images and questions. However, LLMs often struggle in accurately interpreting specific question meanings. Meanwhile, although knowledge graph has rich entity relationships, it is challenging to effectively connect entities to individual image content for visual question answers. In this paper, we propose a novel design to combine knowledge graph and LLMs for zero-shot visual question answer. Our approach uses LLMs' powerful understanding capabilities to accurately interpret image content through a strategic question search mechanism. Meanwhile, the knowledge graph is used to expand and connect users' queries to the image content for better visual question answering. An optimization algorithm is further used to determine the optimal weights for the loss functions derived from different information sources, towards a globally optimal set of candidate answers. Experimental results on two benchmark datasets demonstrate that our model achieves state-of-the-art (SOTA) performance. Both source code and benchmark data will be released for public access. 
\end{abstract}

\begin{document}

\flushbottom
\maketitle
\thispagestyle{empty}
% \date{October 2024}

Visual question answering (VQA) tasks present significant challenges. In these tasks, a model is given an image and a corresponding question, and it must generate an answer based on the information in the image \cite{wu2017visual}. Unlike image captioning, which involves describing visible content, VQA requires the model to interpret and respond to questions that often involve understanding context, reasoning, and inferring details not explicitly present in the image. Nowadays, researchers have developed various approaches to train models for VQA using labeled training data. A typical approach is employing attention mechanism to fuse multimodal features better\cite{sharma2021MedFuseNet,yalin2022MulAtt}. However, although these methods have improved the model's reasoning and generalization abilities to a certain extent, they still necessitate retraining the model from scratch whenever new objects,questions,or answersare are introduced. To address this limitation, Zero-Shot Visual Question Answering (ZS-VQA) has been proposed. ZS-VQA enables models to predict answers about objects, questions, or answers that were not present in the training samples. These methods leverage a wide range of knowledge, from common sense to encyclopedic information about specific elements within the image \cite{antol2015vqa}. For example, incorporating external knowledge, such as Wikipedia \cite{wu2022multi,gui2021kat,lin2022revive} and ConceptNet \cite{liu2004conceptnet,garderes2020conceptbert,marino2019ok}, to enhance the information extraction capability of models, or utilizing knowledge graphs to improve the reasoning ability \cite{wang2015explicit,wang2017fvqa,chen2021zero,wu2023resolving}. However, the open-domain images are vast, and external knowledge sources cannot cover all possible information. Additionally, existing models predominantly focus on understanding the known world and struggle to fully comprehend and reason about images or questions that are not represented in the dataset.

In recent years, large language models (LLMs) like GPT-3 \cite{brown2020language} have demonstrated strong generalization capabilities in natural language processing tasks, such as information extraction and logical reasoning \cite{liang2024toa,kang2024chatmof,cai2024using}, due to their advanced intellectual reasoning abilities \cite{wei2022chain}. However, current LLMs primarily rely on their own internal understanding to resolve ambiguities and perform question reasoning \cite{lan2023improving}. This reliance can introduce unexpected biases, as the models may not fully comprehend the underlying meaning of objects in images. Furthermore, LLMs are less adept at handling uncertain questions and can be easily disrupted by noise in both images and text.

To address the aforementioned limitations, this paper proposes a novel model that combines knowledge graphs and large language models for enhanced zero-shot visual question answering. The model comprises two main components: the answer generation component powered by LLMs, and the answer selection component utilizing knowledge graphs. The LLM component employs an image captioning model to convert images into corresponding textual descriptions and utilizes a question search strategy to diversify the question set appropriately. These image descriptions and questions are then fed into an LLM, which generates a candidate set of answers based on specific semantic representations. The knowledge graph answer selection component models the \textit{[entity, relation, question]} triplets derived from a knowledge graph. It defines three loss functions related to answer generation, entity recognition, and relation identification. This component then refines the candidate answers generated by the LLM by integrating them with the knowledge graph-based candidate set, enhancing the overall accuracy and relevance of the answers.

\begin{figure*}[hbtp]
    \centering
    \includegraphics[width=0.85\linewidth]{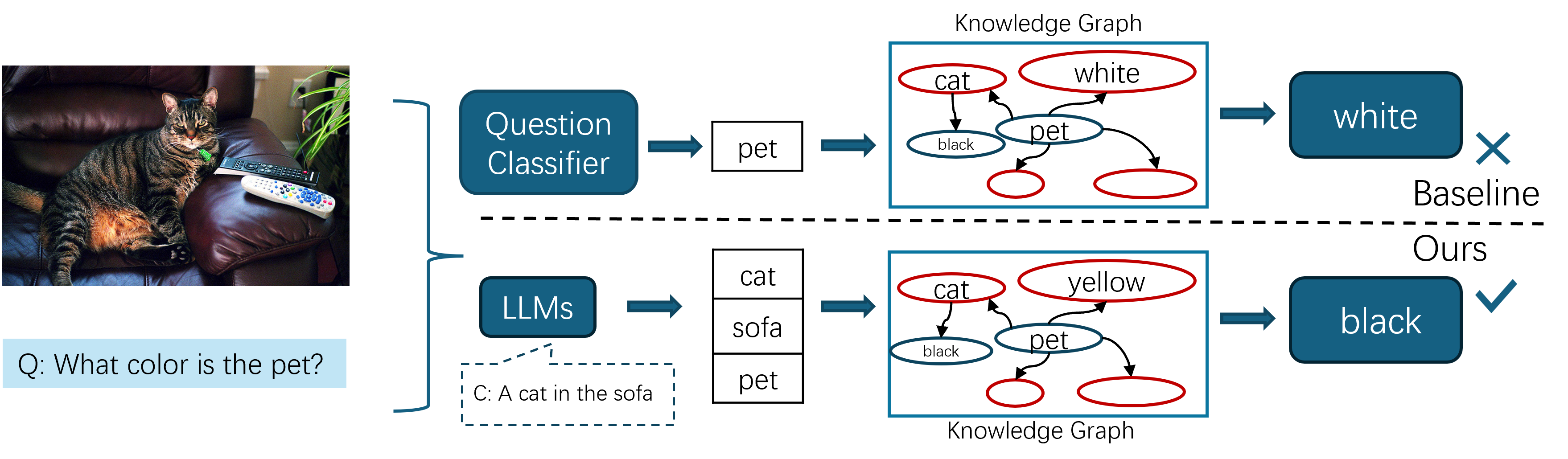}
    \caption{Comparison between existing methods using knowledge graph \textit{vs.} our approach using LLMs and knowledge graph on VQA tasks. It is very common that an image matches to the users' query at the semantic level, but not at the word level. For example, the ``pet'' in the query is actually refer to the ``cat'' in the image. Our approach combines LLMs and knowledge graph to benefit VQA: (1) LLMs precisely interpret the image content, and (2) knowledge graph will help connect users' query to the image content, though their rich entity relationships.}
    \label{LLM}
\end{figure*}

As illustrated in Fig.\ref{LLM}, the baseline method relies solely on a question classifier to establish an initial correspondence between the image-related targets in the question, but it fails to accurately match the entities in the knowledge graph with the targets in the image. In contrast, our method leverages LLMs to precisely extract target entities and identify corresponding answers within the knowledge graph in a more detailed manner. To achieve effective integration and knowledge fusion, Particle Swarm Optimization is employed to calculate the weights for different loss functions, and a scoring function is used to select the optimal ratio for generating the highest-scoring candidate answer set. Overall, our proposed model can produce visual question answers that are more aligned with real-world scenarios and exhibit higher accuracy.

In summary, we make the following contributions:
\begin{itemize}
\item We introduce a novel framework for zero-shot visual question answering that integrates large language models and knowledge graphs effectively.
\item By diversifying the input questions appropriately, we enhance the understanding capability of large language models towards textual inputs, consequently improving the accuracy of answer predictions.
\item Through optimization of the loss function weights associated with both the large language model and knowledge graph, we achieve a more effective integration of the two modules, resulting in a candidate set of answers with increased confidence.
\end{itemize}

\section*{Results}

\begin{table*}[h!]
\centering
\caption{Overall results (\% for Hit@K) on the ZS-F-VQA dataset under the setting of ZSL/GZSL.}
\label{ZSFVQA-compare}
\resizebox{\textwidth}{!}{%
\begin{tabular}{@{}l|ccccc|ccccc@{}}
\toprule
\multicolumn{1}{c|}{} & \multicolumn{5}{c|}{\textbf{GZSL}} & \multicolumn{5}{c}{\textbf{ZSL}} \\ \midrule
\multicolumn{1}{c|}{Methods} & Hit@1$\uparrow$ & Hit@3$\uparrow$ & Hit@10$\uparrow$ & MRR$\uparrow$ & MR$\downarrow$ & Hit@1$\uparrow$ & Hit@3$\uparrow$ & Hit@10$\uparrow$ & MRR$\uparrow$ & MR$\downarrow$ \\ \midrule
Up-down\cite{anderson2018bottom} & 0.00 & \textbf{2.67} & 16.48 & - & - & 13.88 & 25.87 & 45.15 & - & - \\
BAN\cite{kim2018bilinear} & 0.22 & 4.18 & 18.55 & - & - & 13.14 & 26.92 & 46.90 & - & - \\
MLP & 0.07 & 4.07 & 27.40 & - & - & 18.84 & 37.85 & 59.88 & - & - \\
SAN\cite{yang2016stacked} & 0.11 & 6.27 & 31.66 & 0.096 & 48.18 & 20.42 & 37.20 & 62.24 & 0.331 & 19.14 \\
ZS-F-VQA\cite{ramesh2021zero} & 29.39 & 43.71 & 62.17 & 0.401 & 29.52 & 46.87 & 62.00 & 78.14 & 0.572 & 12.22 \\
R-ZS-F-VQA\cite{wu2023resolving} & {\ul \textbf{49.04}} & {\ul \textbf{61.88}} & {\ul 73.61} & {\ul 0.577} & {\ul 23.90} & {\ul 59.39} & {\ul 72.45} & {\ul 82.49} & {\ul 0.676} & {\ul 11.27} \\
Ours & 45.92 & 59.11 & \textbf{74.66} & \textbf{0.581} & \textbf{22.11} & \textbf{60.36} & \textbf{74.11} & \textbf{82.99} & \textbf{0.712} & \textbf{10.96} \\ \bottomrule
\end{tabular}%
}
\end{table*}

\begin{table*}[h!]
\centering
\caption{Comparative results on the F-VQA dataset.}
\label{FVQA-compare}
\begin{tabular}{@{}l|ccccc@{}}
\toprule
 & Hit@1$\uparrow$ & Hit@3$\uparrow$ & Hit@10$\uparrow$ & MRR$\uparrow$ & MR$\downarrow$  \\ \midrule
Up-down\cite{anderson2018bottom} & 34.81 & 50.13 & 64.37 & - & - \\
BAN\cite{kim2018bilinear} & 44.02 & 58.92 & 71.34 & - & - \\
MLP & 34.12 & 52.26 & 69.11 & - & - \\
SAN\cite{yang2016stacked} & 41.62 & 58.17 & 72.69 & 0.605 & 14.75 \\
ZS-F-VQA\cite{ramesh2021zero} & 58.27 & 75.2 & 86.4 & 0.685 & 11.72 \\
R-ZS-F-VQA\cite{wu2023resolving} & {\ul 66.81} & {\ul 80.3} & {\ul 88.66} & - & - \\
F-VQA\cite{wang2017fvqa} & 58.76 & - & - & - & - \\
HQIPV\cite{lu2016hierarchical} & 43.14 & 59.44 & 72.20 & - & - \\
Ours & \textbf{67.52$\pm$ 1.25} & \textbf{79.58$\pm$ 2.12} & \textbf{89.12$\pm$ 1.58} & \textbf{0.699$\pm$ 0.01} & \textbf{10.63$\pm$ 0.57} \\ \bottomrule
\end{tabular}
\end{table*}

\begin{table*}[h!]
\centering
\caption{Ablation experiments on F-VQA.}
\label{ablation}
\begin{tabular}{@{}cccc|ccccc@{}}
\toprule
Baseline & LLM & QS & PSO & Hit@1$\uparrow$ & Hit@3$\uparrow$ & Hit@10$\uparrow$ & MRR$\uparrow$ & MR$\downarrow$  \\ \midrule
\checkmark &  &  &  & 38.64 & 43.71 & 63.14 & 0.401 & 21.55 \\
\checkmark & \checkmark &  &  & 44.50 & 50.12 & 70.89 & 0.450 & 14.62 \\
\checkmark & \checkmark & \checkmark &  & 50.25 & 62.46 & 78.53 & 0.515 & 13.11 \\
\checkmark &  &  & \checkmark & 46.21 & 56..23 & 80.12 & 0.566 & 12.64 \\
\checkmark & \checkmark &  & \checkmark & 59.98 & 74.92 & 82.17 & 0.636 & 11.34 \\
\checkmark & \checkmark & \checkmark & \checkmark & \textbf{62.52} & \textbf{79.58} & \textbf{89.12} & \textbf{0.699 }& \textbf{10.63} \\ \bottomrule
\end{tabular}
\end{table*}

\subsection*{Comparison with the State-of-the-arts}
The performance of each comparative method is measured by evaluating the results of unsupervised image captioning. Table \ref{ZSFVQA-compare} and Table \ref{FVQA-compare} present the VQA results on the F-VQA and ZS-F-VQA datasets, respectively. The best results are indicated in bold. The analysis of the experimental results is as follows:

\textbf{Analysis of the comparative experiments based on ZS-F-VQA:} We selected the ZS-F-VQA dataset and conducted two experimental settings during the testing phase. Table \ref{ZSFVQA-compare} indicates that models designed for traditional VQA tasks, such as Up-Down, are not applicable to the ZS-F-VQA dataset, as the Hit@1 scores are less than 1\%. This suggests that the predicted answers are fundamentally incorrect.

\textbf{Analysis of the comparative experiments based on F-VQA:} Table \ref{FVQA-compare} compares our method with state-of-the-art models based on the F-VQA dataset. In all these settings, the results indicate that our model outperforms the corresponding classifier-based or mapping-based models to varying degrees. The stable performance improvement achieved by our model suggests that incorporating our method into other end-to-end frameworks in the context of generalized knowledge V can also lead to consistent performance improvements.

\subsection*{Ablation study}
We conducted ablation studies to validate each component of our model, shown in Table \ref{ablation}.

\textbf{Effect of LLMs:} The model leverages LLMs by inputting both the generated image description and the question into the LLM, enabling it to generate the corresponding answer. Experimental results demonstrate that, compared to the baseline method, incorporating LLMs into the model significantly enhances performance. This improvement is likely due to the model's ability to better utilize the extensive knowledge base inherent in LLMs, thereby enriching the composition of the answer candidate set. However, since LLMs are designed for language processing and cannot directly handle images, the accuracy of the generated answers relies on the quality of the image description generation. Therefore, it is crucial to impose constraints on the input and output of LLMs to ensure accurate results.

\textbf{Effect of QS:} In this part, we restrict the input to the LLMs by appropriately expanding the question. Specifically, we extract the words most relevant to the image from the question and search for the top-k nearest neighboring words in the entire corpus. We then replace the generated questions based on the overall fluency of the sentence. Finally, these k+1 questions are inputted into the LLM model. Experimental results show that compared to the LLM model alone, the LLM+QS performs better, with the Hit@k metric in the F-VQA dataset improving by an average of 8\%. This further demonstrates the necessity of imposing restrictions on the large language model to achieve better results.

\textbf{Effect of weights optimization:} In this part, due to the existence of two loss functions in the LLM and three loss functions in the knowledge graph, it is not feasible to determine the values of $(\lambda_1, \lambda_2, \lambda_3, \lambda_4, \lambda_5)$ solely through parameter tuning. Therefore, we use the Particle Swarm Optimization (PSO) method to calculate the weights of different losses. By evaluating the scoring function of the answer candidate set, we search for the optimal weights that best match the image and question. Experimental results show that using PSO for optimization alone improves the model by an average of 10 percentage points in the Hit@k metric compared to the baseline.

\begin{figure*}[t]
  \centering
  \includegraphics[scale = 0.34]{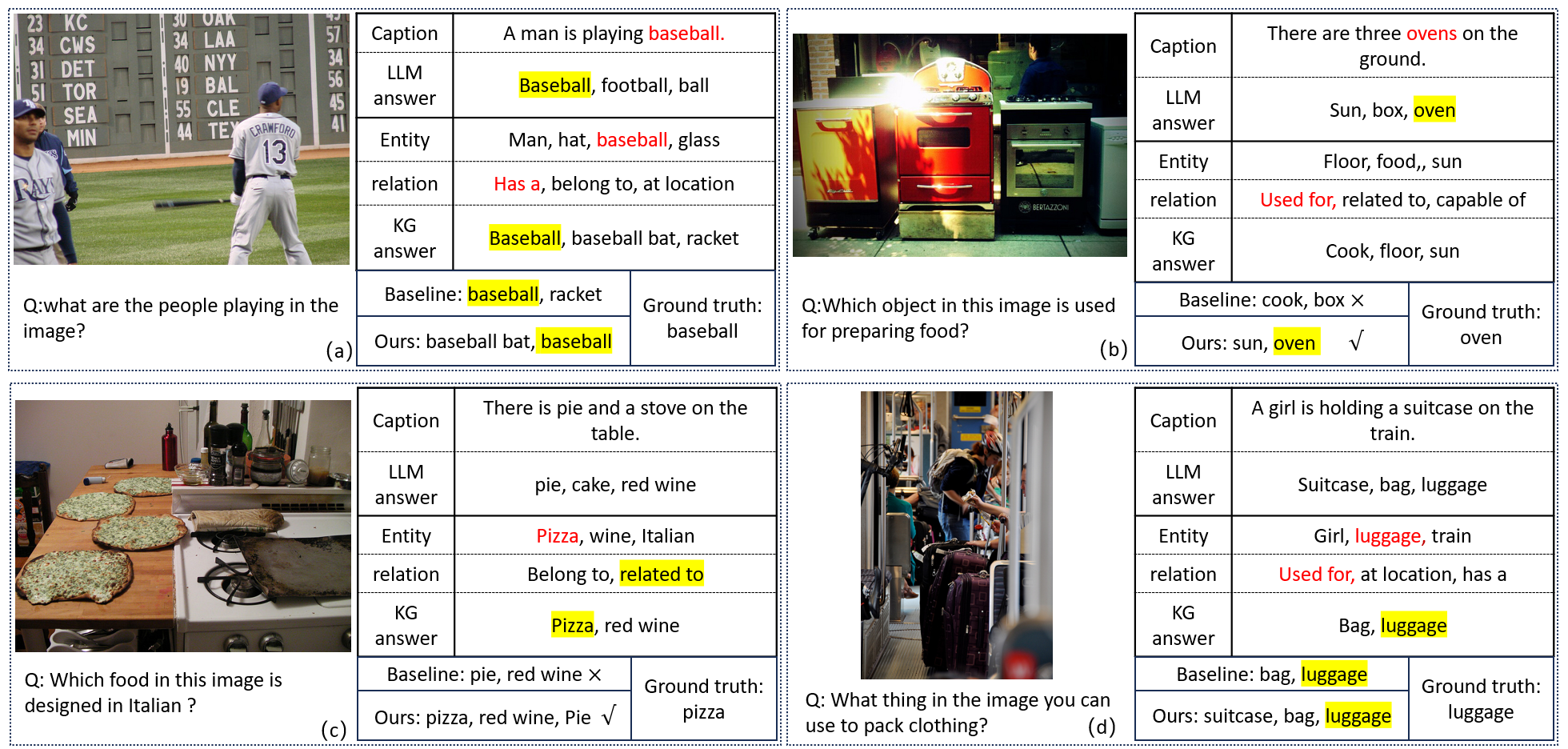}
  \caption{Typical examples of VQA generated by our model.}
  \label{qua-fig}
\end{figure*}

\subsection*{Parameter sensitivity analysis}
The parameter sensitivity analysis experiments were conducted to analyze two hyper-parameters: the search threshold $\mu$ for question generation and the number of iterations k in the PSO.

\textbf{Question search threshold $\mu$:} As shown in Table \ref{sen-mu}, we controlled the search threshold $\mu$ for question generation to determine which similar questions could be selected into the question candidate set. 
% If the search threshold is too small, almost all questions will be used as input to the LLMs, which can lead to overfitting and prevent LLMs from truly understanding the content of the image. On the other hand, if the search threshold is too large, only a few questions will be selected into the question candidate set, reducing the diversity of the questions and negatively impacting the model's performance.
Through parameter analysis, we found that the model performs best when the search threshold for questions is set to 0.7.

\begin{table}[h!]
\centering
\caption{Sensitivity analysis on the search threshold $\mu$.}
\label{sen-mu}
  \scalebox{0.85}{
\begin{tabular}{@{}c|ccc|ccc@{}}
\toprule
 & \multicolumn{3}{c|}{\textbf{ZSL}} & \multicolumn{3}{c}{\textbf{GZSL}} \\ \midrule
$\mu$ & Hit@1& Hit@3& Hit@10& Hit@1& Hit@3& Hit@10 \\\midrule
0.2& 34.81& 50.13& 64.37& 33.81& 49.13& 70.42 \\
0.4& 44.02& 58.92& 71.34& 41.42& 56.82& 69.85 \\
0.7& \textbf{58.27}& \textbf{75.2}& \textbf{86.4}& \textbf{45.92}& \textbf{59.11}& \textbf{70.66} \\
0.8& 41.62& 58.17& 72.69& 41.82& 58.77& 69.22 \\
0.9& 34.12& 52.26& 69.11& 34.57& 52.23& 69.14 \\ \bottomrule
\end{tabular}}
\end{table}

\begin{figure}[h!]
  \centering
  \includegraphics[scale = 0.53]{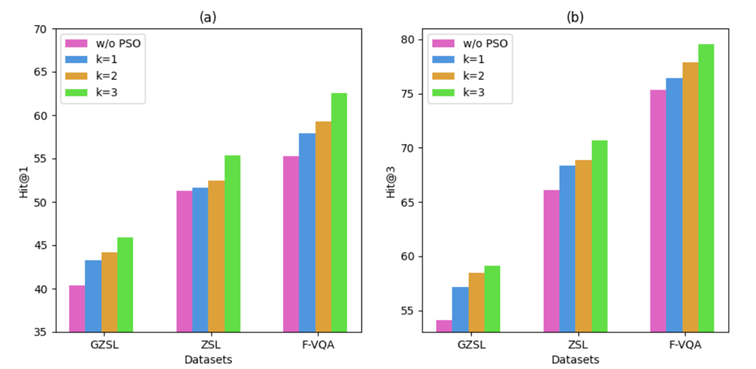}
  \caption{Sensitivity analysis for the hyperparameter k.}
  \label{sen-k}
\end{figure}

\textbf{The number of iterations k in PSO:} By designing the number of iterations k in the Particle Swarm Optimization (PSO) algorithm, we primarily use it when the score function S is continuously less than the optimal score for K times.  Fig.\ref{sen-k} illustrates the effect of selecting different thresholds k on the model for different datasets. When $k=0$, it indicates that the model does not use PSO evolutionary computation. The results show that when $k=3$ better performance can be achieved.

\subsection*{Qualitative results}
In this section, we compared VQA using different methods and visualized their performance, as shown in Fig.\ref{qua-fig} We selected ZS-F-VQA as the VQA dataset and generated image descriptions using an image description method tailored for VQA. Unlike other methods, our answer candidate set includes answers generated by the LLMs and those generated based on the knowledge graph. The final answer candidate set consists of the union of LLM-generated answers and knowledge graph answers. From the figure, it can be observed that the answers generated by LLMs rely heavily on the image description, which inevitably depends on the pre-trained model for image description. Conversely, by using the knowledge graph to extract entities and relationships relevant to the image and question, the model can accurately reflect the relevance between the current image and the question. This dual approach ensured a more comprehensive and accurate generation of answers.

\section*{Discussion}
\subsection*{VQA based on external knowledge} %related work
% While traditional methods can effectively extract the content of images and answer related questions, there are instances where external knowledge is required to answer corresponding questions, such as common sense reasoning or task-specific knowledge or encyclopedia-style information. To address this, researchers have connected VQA models with external knowledge repositories.

Wang et al. \cite{wang2015explicit} initially identified relevant concepts from images by integrating image data with knowledge base concepts, matching them with semantic concepts in the knowledge base to generate queries for natural language questions. However, this approach did not address the issue of answer bias—many answers might not have been encountered during training (i.e., unseen answers). To improve upon this, Wang et al. \cite{wang2017fvqa} introduced the FVQA method. This method employs LSTM and image-question mapping to pinpoint significant content in images and generate queries based on both the image and the knowledge base. Chen et al. \cite{chen2021zero} introduced a new answer-based zero-shot VQA segmentation dataset, ZS-VQA, designed for the F-VQA dataset. Most recently, Wu et al. \cite{wu2023resolving} focused on enhancing the accuracy of extracting entities and relationships from the knowledge base. They achieved this by employing a specialized pre-training model for representing input information and utilizing a contrastive learning-based common feature space for information retrieval. Similarly, Fei et al. \cite{fei2022brivl} developed a large-scale multimodal foundation model utilizing cross-modal contrastive learning. They demonstrated that the weak semantic correlation train data helped improve the generality and cognition of the model to perform VQA task.

\subsection*{VQA based on LLMs} %related work
Unlike traditional VQA training tasks, the VQA paradigm utilizing Large Language Models (LLMs) directly relies on natural language as an intermediate representation of images, thus eliminating the need for expensive pre-training. Liang et al. \cite{liang2024toa} introduced TOA, where LLMs make an initial hypothesis based on their knowledge and then actively gather the visual evidence needed to verify this hypothesis. Guo et al. \cite{guo2023images} proposed Img2LLM, a plug-and-play module that converts images into synthesized question-answer pairs derived solely from the current question image. Additionally, Lan et al. \cite{lan2023improving} developed a question-guided visual question answering reasoning model to consider sentence fluency, semantic completeness, and syntactic invariance. Other researchers have explored the task planning capabilities of LLMs. For example, Gupta et al. \cite{gupta2023visual} and Surís et al. \cite{suris2023vipergpt} suggest that LLMs can generate programs for subtasks executed by predefined sub-modules. This approach is supported by works such as those by Khot et al. \cite{khot2022decomposed}, Huang et al. \cite{huang2022inner}, and Wang et al. \cite{wang2023describe}, who have shown the effectiveness of decomposing complex tasks into manageable subtasks that LLMs can handle efficiently.

\subsection*{Conclusion}
In this paper, we propose a novel method that combines a knowledge graph with a LLMs for enhanced zero-shot visual question answering. To address the traditional large language models' lack of sensitivity to noise and their inability to accurately comprehend the specific meaning of a question, our model consists of two main components: the knowledge graph for answer aggregation and the LLMs for answer generation. The knowledge graph extracts entities and relationships from images and questions, while the LLMs refines these entities, enabling more accurate image recognition and question answering. Experimental results demonstrate that our model surpasses state-of-the-art methods on benchmark datasets.

\subsection*{Limitation}
Although the proposed evolutionary computation-based zero-shot visual question answering in this paper effectively enhances the model's ability to understand external knowledge by leveraging large language models, it still requires other tools to bridge the gap between images and text. Therefore, effectively transferring image features to large language models like image captioning, will be an important direction for future improvements in the knowledge question answering model proposed in this paper.In this paper, we only utilized OPT as the large language model. However, employing the latest GPT-4 model would potentially yield higher performance for the model.

\subsection*{Societal impacts}
In real-life scenarios, leveraging visual question answering models can greatly assist individuals with visual impairments in enhancing their understanding of the natural environment. Additionally, it can provide them with additional learning resources and tools to better comprehend the world around them. Furthermore, due to the significant cost of manually annotating large-scale image-text paired datasets required for training large models, zero-shot visual question answering can effectively provide annotated datasets for multimodal large models.However, if a large number of images are used in visual question answering models, it may lead to copyright infringement issues and the potential for malicious tampering.

\begin{figure*}[h!]
    \centering
    \includegraphics[width=0.85\linewidth]{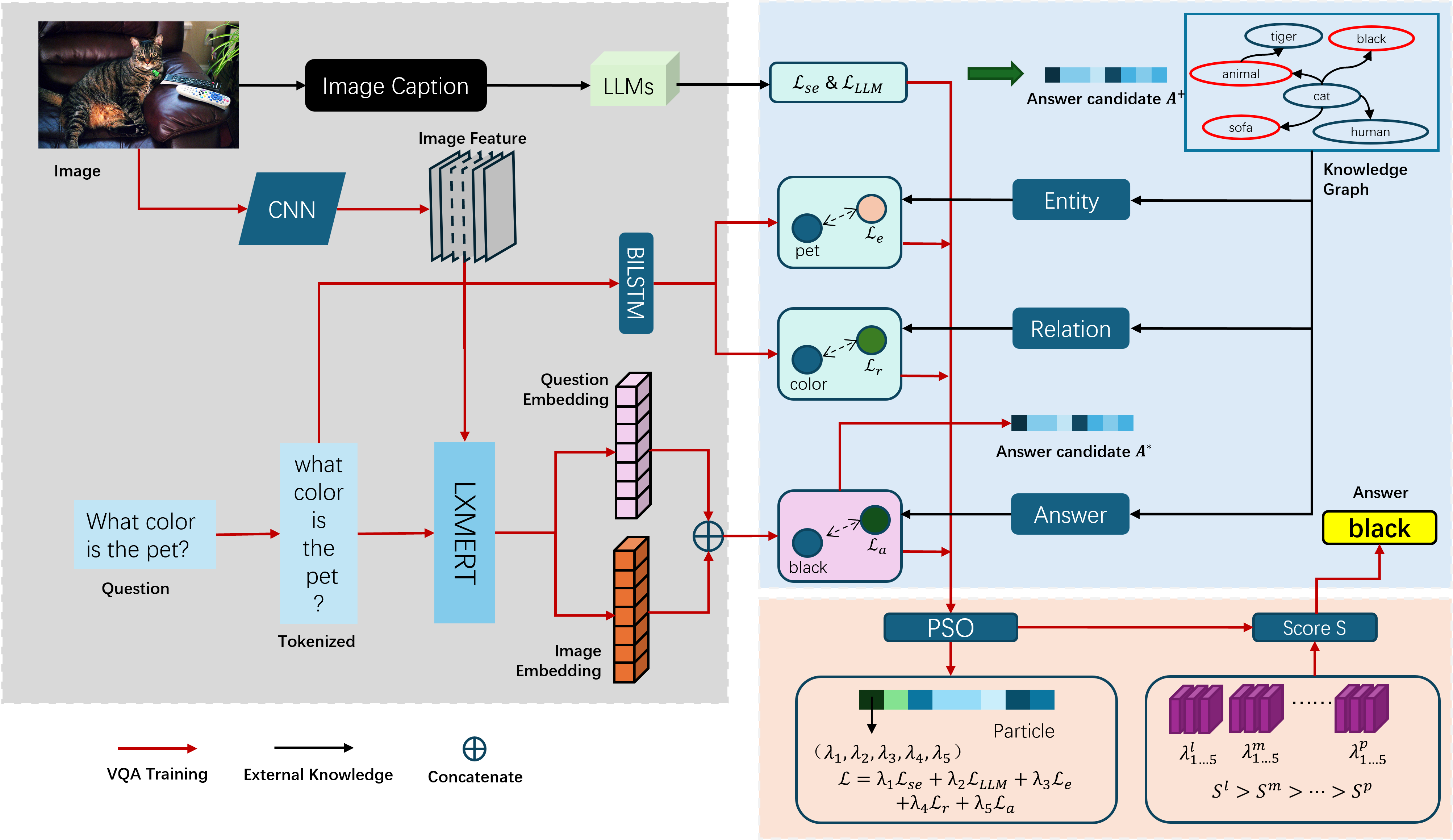}
    \caption{Overview of our proposed model.The model includes VQA training module, external knowledge module (LLMs and knowledge graph) and PSO training module.}
    \label{model}
\end{figure*}

\section*{Methods}
To address the issue of noise in generating answers with traditional large language models, we have developed a joint framework that combines a large language model with a knowledge graph to generate a set of answer candidates, as illustrated in Fig.\ref{model} To further improve the model's granularity in image recognition, we utilized the particle swarm optimization algorithm to evaluate the influence of various external knowledge sources on the answers.

\subsection*{Answer generation through large language model and question search strategy}
Since LLMs are inherently designed for natural language processing, they cannot directly process image data as input. Therefore, it is necessary to establish a connection between images and text. Inspired by Zhang et al.\cite{zhang2021vinvl}, we use VinVL as a pre-trained model for image caption generation, which produces informative descriptions of the current image. These descriptions are then used as supplementary inputs to the LLMs. However, due to the extensive information contained in images, directly generating descriptions can result in sentences that may not be relevant to the question. To address this issue, we tokenize each word of the question and input them alongside the image vector:

\begin{equation}
    (O,\mathbf{f}_i )=\textit{Faster-RCNN}(I_i ) \;\; i=1,…,k
\end{equation}
\begin{equation}
    Y=\textit{VinVL}((Q,O),\mathbf{f}_i )
\end{equation}

where $k$ represents the number of images, $\mathbf{f}_{i}\in\mathbb{R}^{m}$ denotes the distribution representation in the high-dimensional latent space obtained from the Faster-RCNN model, $O$ represents the detected objects, $Q$ represents the question vector associated with the current image to be answered, and $Y$ represents the generated image description. To ensure that each image has a unique caption, we incorporate visual concepts generated by a convolutional neural network as part of the question and input them together with the image features into the pre-trained model for generating image descriptions. Additionally, we repeat this process $M$ times, filter out duplicate sentences, and obtain $k$ distinct and relevant captions for the images.

However, in LLMs, there are instances where the model may fail to fully grasp the true meaning of an object in an image. Additionally, LLMs exhibit low sensitivity to ambiguous questions and can be easily disrupted by noise in the images or text. This means that the model may not accurately identify which region of the image corresponds to the question or the specific intent of the question. To address this, we employ a Question Search (QS) strategy. In this strategy, keywords in the question are replaced with various alternatives to enable the LLMs to develop a more comprehensive understanding of the question. To ensure that irrelevant words are not replaced, we calculate the similarity between each word in the image description and the objects present in the image, allowing us to distinguish them:
\begin{equation}
     \mathbf{P}(q)=\arg\max(\text{sim}(q,o_i )) \;\;  i=1,…,m
\end{equation}
where $m$ represents the number of targets extracted from each image, $sim$ function denotes the cosine similarity between words, and q represents the generated word. We set a threshold $\mu$, and if P(q) >$\mu$, we search for the top-k neighboring words of the current word for diverse replacements. To prevent changes in sentence structure and semantics, we ensure that only one word in a question is replaced:
\begin{equation}
  \hat{q}={q, q_1, q_2, ..., q_k} \;\; \text{if} \;\; p(q)>\mu\ \;\; \text{and} \;\; \text{sim}(q_i,q)>\delta
\end{equation}
where $q_i$ represents the neighboring word, and $\delta $represents the similarity interval. Therefore, we can utilize the generated question and the original question to construct a loss function $\mathcal{L}_{se}$ to measure the diversity of the generated questions and assess their quality:
\begin{equation}
    \mathcal{L}_\mathrm{se}=-\sum_{i=1}^m\sum_{j=1}^lP(q_j)p(Q_i)\log\frac{p(Q_i)}{q(Q_i)}
\end{equation}
where $m$ represents the number of questions, $P(q_j)$ denotes the probability of the j-$th$ word being replaced in the i-$th$ question, $p(Q_i)$ represents the distribution of the original question for the i-$th$ image, and $q(Q_i)$ represents the distribution of the generated target question for the i-$th$ image.

We sequentially input the $k$-question prompts into the LLM and perform greedy decoding for each prompt. The LLM is tasked with outputting a set of answer candidates along with their confidence scores. By inputting various image descriptions and their corresponding questions into the frozen LLM, we obtain the respective answers and their associated probabilities as follows:
\begin{equation}
    \left(\hat{\mathrm{A}}_i,\mathrm{P}(\hat{\mathrm{A}}_i)\right)=LLM(Q_i,Y_i) \;\; i=1,...,n
\end{equation}
where $n$ represents the number of generated answers. To ensure the fluency of the generated answers in terms of sentence structure, the above model incorporates a probability language model (LM) to assess whether the generated answers adhere to grammatical and logical rules. By training a joint probability scoring function, the aim is to maximize the likelihood of generating sentences. This can be expressed as follows:
\begin{equation}
    {F}_{\mathrm{LM}}(\hat{A})=\ln{P}_{\mathrm{LM}}(\hat{A})=\ln\prod_{i=1}^TP(w_i|w_{i-1},...,w_1)
\end{equation}
where $w_i$ represents the i-$th$ word of the answer $\hat{A}$, and $T$ represents the length of the question. This approach allows us to obtain the scoring function for the answer $\hat{A}$. Utilizing the joint probability $F_{LM}(\hat{A})$ and the probability of the currently generated answer, we then construct the loss function for the LLM:
\begin{equation}
    \mathcal{L}_{\mathrm{LLM}}=\sum_{i=1}^m\mathrm{P}(\hat{\mathrm{A}}_\mathrm{i})\mathrm{~F}_{\mathrm{LM}}(\hat{\mathrm{A}}_\mathrm{i})
\end{equation}

\subsection*{Enhanced answering through knowledge graph}
In VQA, the knowledge graph can be represented as a triplet $G = [E, R, A]$, where E denotes the entities, R denotes the relationships between entities, and A denotes the answers. The knowledge graph provides relationships and entities that can be used to compare corresponding components in images and sentences, as well as to provide answer candidates. Since question Q itself includes entities and relationships, we utilize Bi-LSTM to extract contextual information from question Q, resulting in a representation of the question, denoted as $Q$. This allows us to better identify the specific location of entities and relationships in the question. Furthermore, because the dimensions of the entity set, relationship set, and question representation in the knowledge graph are different, we need to align the dimensions. This is achieved using a forward propagation network consisting of fully connected layers and regularization. The process can be represented using the following formulas:
\begin{equation}
    Q=BiLSTM(w_i )
\end{equation}
\begin{equation}
    R(y)=FN(y)
\end{equation}
where y represents any set from the entity set, relationship set, and question set. This way, we obtain entity and relationship representations from the knowledge graph and question in the same dimension. We aim to train the model to ensure that the entities and relationships from the knowledge graph can be paired one-to-one with the corresponding sets in the question. Therefore, we use a commonly used loss function in contrastive learning, InfoNCE loss\cite{he2020momentum}, as follows:
\begin{equation}
    \mathcal{L}_\mathrm{e}=-log\frac{\exp{(\mathbb{Q}\cdot R(e)_+/\tau)}}{\sum_{i=1}^k\exp{(\mathbb{Q} R(e)_i/\tau)}}
\end{equation}
\begin{equation}
    \mathcal{L}_{\mathrm{r}}=-log\frac{\exp\left(\mathbb{Q}\cdot R(r)_{+}/\tau\right)}{\sum_{i=1}^{k}\exp\left(\mathbb{Q} R(r)_{i}/\tau\right)}
\end{equation}
where $L_e$ and $L_r$ represent the loss functions for entities and relationships, respectively, and $\tau$ is a hyper-parameter. However, the answer set in the knowledge graph has not been aligned with features. Therefore, we utilize a pre-trained multimodal model such as LXMERT \cite{tan2019lxmert} to fuse the question features and image features, obtaining a shared feature $F_iq$ for the question and image:
\begin{equation}
    \mathrm{F_{iq}}=LXMERT(F_{img},w_i)
\end{equation}

Then, we pair this feature with the dimension-transformed answer set and obtain the corresponding loss function:
\begin{equation}
    \mathcal{L}_\mathrm{a}=-log\frac{\exp\left(F_{iq}\cdot R(a)_+/\tau\right)}{\sum_{i=1}^k\exp\left(F_{iq}\cdot R(a)_i/\tau\right)}
\end{equation}

In this way, the model obtains the loss function for the answer set. At the same time, the initial candidate set for answers can be obtained using the following:
\begin{equation}
    \hat{a}=R(a)^TF_{iq}
\end{equation}

Now that we have obtained the loss functions $\mathcal{L}_\mathrm{se}$ and $\mathcal{L}_\mathrm{LLM}$ for generating answers using the LLM, as well as the loss functions $\mathcal{L}_\mathrm{e}$, $\mathcal{L}_\mathrm{r}$, and $\mathcal{L}_\mathrm{a}$ for generating question, entity, and answer based on the knowledge graph. The final loss function of the model can be expressed as:
\begin{equation}
\mathcal{L}=\lambda_1\mathcal{L}_{se}+\lambda_2\mathcal{L}_{LLM}+\lambda_3\mathcal{L}_e+\lambda_4\mathcal{L}_r+\lambda_5\mathcal{L}_a
\end{equation}
where $\lambda$ are the weights and will be optimized. See Appendix A.

% \section*{EXPERIMENTAL SETUP}
% Here we describe the datasets regarding ZS-VQA, evaluation metrics, baseline models for reference, and settings for experimental details.

\subsection*{Dataset}
In this paper, we utilized two primary datasets to evaluate our model's performance. The first dataset, the supervised F-VQA dataset, is designed for supervised visual question answering. It consists of 2,190 images, 5,286 questions, and a knowledge graph containing 193,449 facts. Each triplet [image, question, answer] in this dataset is constructed using information from various public knowledge bases, requiring external knowledge for effective answering. The second dataset, the ZS-F-VQA dataset, is used for zero-shot visual question answering. Derived from the F-VQA dataset, it is divided into five parts with no overlap between the answers in the training and test sets. This careful partitioning allows for the creation of zero-shot and fact-based evaluation scenarios, ensuring that the model's ability to generalize to unseen questions and answers is rigorously tested.

\subsection*{Evaluation}
We followed the settings of ZS-F-VQA and evaluated the performance of the models based on the comparison results using the Hit@1, Hit@3, and Hit@10 metrics. Additionally, in the ablation experiments, we used Mean Reciprocal Rank (MRR) and Mean Rank (MR) as additional evaluation metrics. Here, the Hit@k metric is used to indicate whether the ground truth value is ranked within the top k predicted answers.

\subsection*{Implementation details}
The model utilizes a pre-trained ResNet-152 \cite{he2016deep} on ImageNet \cite{deng2009imagenet} to extract visual features. Additionally, a Faster-RCNN \cite{ren2015faster} model trained on the COCO dataset \cite{lin2014microsoft} is used to obtain object and visual features from the images. For the feature extraction of questions and answers, Glove 
\footnote{\url{https://nlp.stanford.edu/projects/glove/}}
vectors are employed to transform textual features into 300-dimensional embeddings, which are then input into a Bi-LSTM \cite{huang2015bidirectional}. As for the large language model, OPT \cite{zhang2023opt} is chosen with parameters of 2 billion. Adam optimizer and progressive learning rate warm-up strategy are used for all models 
\footnote{\url{https://github.com/huggingface/transformers/blob/v4.19.0/src/transformers/models/opt}}
. The learning rate for the first five epochs is set as $(epoch + 1)*10^{-3}$, and thereafter, the decay rate is set to 0.7 for subsequent epochs. The parameter t in the infoNCE loss function is set to 0.01. In order to conduct comparative experiments, we followed the experimental settings of the model \cite{ramesh2021zero} and used standard dataset configurations 
\footnote{\url{https://github.com/China-UK-ZSL/ZS-F-VQA}}
. This includes 5 data splits based on images, and for the experiment, we specified the top 500 candidate answers which accounted for 94.30\% of the total. We use 2*RTX TITAN as our GPU, memory requirement is 48GB.

Due to the limitations of traditional VQA datasets in comprehending information from images and questions and adequately handling answers not appearing in the training samples, we studied two configurations for the testing phase of ZSVQA: ZSL and GZSL. In the ZSL setting, the answer candidate set in the test samples (i, q, a) consists of answers not present in the training set, i.e., the unseen dataset $A_u$. In the GZSL setting, the answer candidate set includes all answers from the training and test sets, i.e.,$ A_u $and$ A_s$, where $ A_u\cap A_s = \emptyset$. 

% supplemental material
\subsection*{Particle Swarm Optimization to find the optimal weights}
Selecting the weights for the corresponding loss functions is an important consideration for the model. If we directly train the model by freezing the weights of the loss functions, it is difficult to obtain the optimal solution, and the generated answer candidates may in turn affect the weights of the corresponding loss functions, ultimately impacting the accuracy of the answers. To address this, we designed a combination of multiple loss functions as the training objective. During training, we utilize the Particle Swarm Optimization (PSO) algorithm to select an adaptively weighted combination of loss functions. In each training iteration, the PSO algorithm is employed to calculate the optimal weights through evolutionary computation. The PSO requires a scoring function to control the quality of the currently generated answers.

Firstly, for the knowledge graph, we can set a similarity score to measure the similarity between the question and entities/relationships.
\begin{equation}
    \mathrm{sim}(\mathrm{x})=\frac{R(\hat{y})^T\mathbb{Q}}{|R(\hat{y})||Q|}
\end{equation}
Then, we can calculate the similarity scores for relationships and entities as a constraint to compute the candidate set for answers based on the knowledge graph.
\begin{equation}
    {A}^*=\{a|sim(\hat{e})+sim(\hat{r})>\delta\wedge(\hat{e},\hat{a},\hat{r})\in G\wedge\hat{e}\in E\wedge\hat{r}\in R\}
\end{equation}
where $\hat{e}$ and $\hat{r}$ represent the entities and relationships mentioned in the question, $\hat{a}$ represents the initial candidate answer set, $sim(\hat{e})$ and $sim\hat{r})$ represent the similarity scores for entities and relationships, E represents the entity set, R represents the relationship set, and G represents the set of triplets that includes the three sets. $\delta$ is the threshold used to control the similarity score. Additionally, since the large language model (LLM) also generates corresponding answer candidate sets, we combine the candidate sets from the knowledge graph and LLM to form the final answer candidate set.

Therefore, the final scoring function can be determined by both the predictions from the LLMs and the answers computed from the knowledge graph. If the predicted answer is in the answer candidate set $A^+$,where ${A}^+=A^*\cup\hat{A}$ we increase the score; otherwise, we decrease the score. The scoring function for the predictions from the LLMs can be defined as follows:
\begin{equation}
    S_{LLM}=\lambda_1P(a)f_{LM}(a)
\end{equation}
Since the newly generated questions are already involved in the answer generation process of the large language models (LLMs), and the loss functions $\mathcal{L}_{LLM}$ and $\mathcal{L}_{se}$ are not completely independent, we did not select a separate scoring function for question search.
\begin{equation}
    S_{G}=R(a)^{T}F_{iq}+\beta(sim(e)+sim(r))
\end{equation}
Since the generated answers can come from both the predictions of the large language models (LLMs), denoted as $A^*$, and the predictions of the knowledge graph, denoted as $A^*$, the overall scoring of the model needs to be discussed based on different scenarios:
\begin{equation}
    \text{score(a)}=\begin{cases}\quad  S_{LLM} \quad\quad a\in\hat{A}\wedge a\notin A^{*}\\\  \quad  S_{G}\quad  \quad \;\;\;\;   a\notin\hat{A}\wedge a\in A^{*}\\ S_{LLM}+S_{G}\;\   a\in\hat{A}\wedge a\in A^{*}\\ R(a)^{T}F_{iq}-b\quad \quad other\end{cases}
\end{equation}
where b is a fixed value used to subtract an appropriate score from predicted answers that do not belong to the answer set. During training, we evaluate the performance of each iteration by scoring the answer candidates formed. If the value of the scoring function remains continuously below the maximum value for k consecutive times, we consider that the current training phase is not progressing towards a better solution. In such cases, PSO is activated to search for the optimal objective function:
\begin{equation}
    S_{best} = max(score(a),S_{best})
\end{equation}

\IncMargin{1em}
\begin{algorithm*}[h!]
    \caption{PSO-VQA for finding the optimal weights}
    \label{pso-vqa}
    \SetKwData{Left}{left}
    \SetKwData{This}{this}
    \SetKwData{Up}{up} 
    \SetKwFunction{Union}{Union}
    \SetKwFunction{FindCompress}{FindCompress} 
    \SetKwInOut{Input}{input}
    \SetKwInOut{Output}{output}
	
	\Input{Image feature $F_{img}$, question embedding $Q$, batch, epoch, particle\_num, iteration, knowledge graph $[e,r,a]$} 
	\Output{The optimized answer candidate set and the corresponding weight parameters $\lambda_i$, $i=1,...,5$}
	 \BlankLine 
	 
    \emph{Initializing weight parameters $(\lambda_1,\lambda_2,\lambda_3,\lambda_4,\lambda_5)$} and $\lambda_1+\lambda_2+\lambda_3+\lambda_4+\lambda_5=1$\; 
    \For{$i\leftarrow 1$ \KwTo $epoch$}{ 
        \For{$v\leftarrow 1$ \KwTo $batch$}{ 
            $\mathcal{L}_e\leftarrow\mathcal{L}(Q,e)$\;
            $\mathcal{L}_r\leftarrow\mathcal{L}(Q,r)$\;
            $\mathcal{L}_e\leftarrow\mathcal{L}(Q,e,F_{img})$\;
        }
        $A^+\leftarrow update(A^+, \mathcal{L}(\lambda_1,\lambda_2,\lambda_3,\lambda_4,\lambda_5))$\;
        $S_{LLM}, S_G \leftarrow evaluate(r,e,a)$\;
        $score(a)\leftarrow (S_{LLM}, S_G)$\;          
        $S_{best}\leftarrow \max(S_{best}, score(a))$\;
        
        \If{$score(a) < S_{best}$ for $K$ times continuously}{
            \For{$j\leftarrow 1$ \KwTo $iteration$}{
                \For{$h\leftarrow 1$ \KwTo particle\_num}{
                    training with PSO\;
                    $A^+ \leftarrow update(A^+, \mathcal{L}(\lambda_{1}^{j,h}, \lambda_{2}^{j,h}, \lambda_{3}^{j,h}, \lambda_{4}^{j,h}, \lambda_{5}^{j,h}))$\;  % 括号已闭合
                    $S^{j,h}\leftarrow score(a)^{j,h}$\;
                }
            }
        }
        $\{S^{j_1,h_1},S^{j_2,h_2},...\}\leftarrow sort(\{S^{j,h}\})$\;
        $(\lambda_1,\lambda_2,\lambda_3,\lambda_4,\lambda_5)\leftarrow (\lambda_{1}^{j,h},\lambda_{2}^{j,h},\lambda_{3}^{j,h},\lambda_{4}^{j,h},\lambda_{5}^{j,h})$\;
    }
\end{algorithm*}
\DecMargin{1em}

After that, we utilize a fitness function to score the weight combinations of each particle during training, and the best result is saved at the end of the PSO training. When it comes to evaluating quality, the higher the prediction on generated data, the better the quality. It measures the gap between the generated samples and the real samples.

\section*{Data availability}
Data related to this paper can be downloaded from: \url{https://github.com/China-UK-ZSL/ZS-F-VQA },  \url{https://github.com/wangpengnorman/FVQA?tab=readme-ov-file}.

\section*{Code availability}
The code to reproduce the experiments is available at \url{https://github.com/Alexlabcode/Fan}.

\newpage

\bibliography{ref}

\begin{thebibliography}{10}
\urlstyle{rm}
\expandafter\ifx\csname url\endcsname\relax
  \def\url#1{\texttt{#1}}\fi
\expandafter\ifx\csname urlprefix\endcsname\relax\def\urlprefix{URL }\fi
\expandafter\ifx\csname doiprefix\endcsname\relax\def\doiprefix{DOI: }\fi
\providecommand{\bibinfo}[2]{#2}
\providecommand{\eprint}[2][]{\url{#2}}

\bibitem{wu2017visual}
\bibinfo{author}{Wu, Q.} \emph{et~al.}
\newblock \bibinfo{journal}{\bibinfo{title}{Visual question answering: A survey of methods and datasets}}.
\newblock {\emph{\JournalTitle{Computer Vision and Image Understanding}}} \textbf{\bibinfo{volume}{163}}, \bibinfo{pages}{21--40} (\bibinfo{year}{2017}).

\bibitem{sharma2021MedFuseNet}
\bibinfo{author}{Dhruv, S.}, \bibinfo{author}{Sanjay, P.} \& \bibinfo{author}{Chandan, R., K.}
\newblock \bibinfo{journal}{\bibinfo{title}{Medfusenet: An attention-based multimodal deep learning model for visual question answering in the medical domain}}.
\newblock {\emph{\JournalTitle{Scientific Reports}}}  (\bibinfo{year}{2021}).

\bibitem{yalin2022MulAtt}
\bibinfo{author}{Yalin, M.}, \bibinfo{author}{Shuyun, H.}, \bibinfo{author}{WenFang, C.}, \bibinfo{author}{Guodong, L.} \& \bibinfo{author}{Meng, T.}
\newblock \bibinfo{journal}{\bibinfo{title}{Research on visual question answering based on dynamic memory network model of multiple attention mechanisms}}.
\newblock {\emph{\JournalTitle{Scientific Reports}}}  (\bibinfo{year}{2022}).

\bibitem{antol2015vqa}
\bibinfo{author}{Antol, S.} \emph{et~al.}
\newblock \bibinfo{title}{Vqa: Visual question answering}.
\newblock In \emph{\bibinfo{booktitle}{Proceedings of the IEEE international conference on computer vision}}, \bibinfo{pages}{2425--2433} (\bibinfo{year}{2015}).

\bibitem{wu2022multi}
\bibinfo{author}{Wu, J.}, \bibinfo{author}{Lu, J.}, \bibinfo{author}{Sabharwal, A.} \& \bibinfo{author}{Mottaghi, R.}
\newblock \bibinfo{title}{Multi-modal answer validation for knowledge-based vqa}.
\newblock In \emph{\bibinfo{booktitle}{Proceedings of the AAAI conference on artificial intelligence}}, vol.~\bibinfo{volume}{36}, \bibinfo{pages}{2712--2721} (\bibinfo{year}{2022}).

\bibitem{gui2021kat}
\bibinfo{author}{Gui, L.} \emph{et~al.}
\newblock \bibinfo{title}{Kat: A knowledge augmented transformer for vision-and-language}.
\newblock In \emph{\bibinfo{booktitle}{Proceedings of the 2022 Conference of the North American Chapter of the Association for Computational Linguistics: Human Language Technologies}}, \bibinfo{pages}{956--968} (\bibinfo{year}{2022}).

\bibitem{lin2022revive}
\bibinfo{author}{Lin, Y.} \emph{et~al.}
\newblock \bibinfo{journal}{\bibinfo{title}{Revive: Regional visual representation matters in knowledge-based visual question answering}}.
\newblock {\emph{\JournalTitle{Advances in Neural Information Processing Systems}}} \textbf{\bibinfo{volume}{35}}, \bibinfo{pages}{10560--10571} (\bibinfo{year}{2022}).

\bibitem{liu2004conceptnet}
\bibinfo{author}{Liu, H.} \& \bibinfo{author}{Singh, P.}
\newblock \bibinfo{journal}{\bibinfo{title}{Conceptnet—a practical commonsense reasoning tool-kit}}.
\newblock {\emph{\JournalTitle{BT technology journal}}} \textbf{\bibinfo{volume}{22}}, \bibinfo{pages}{211--226} (\bibinfo{year}{2004}).

\bibitem{garderes2020conceptbert}
\bibinfo{author}{Gard{\`e}res, F.}, \bibinfo{author}{Ziaeefard, M.}, \bibinfo{author}{Abeloos, B.} \& \bibinfo{author}{Lecue, F.}
\newblock \bibinfo{title}{Conceptbert: Concept-aware representation for visual question answering}.
\newblock In \emph{\bibinfo{booktitle}{Findings of the Association for Computational Linguistics: EMNLP 2020}}, \bibinfo{pages}{489--498} (\bibinfo{year}{2020}).

\bibitem{marino2019ok}
\bibinfo{author}{Marino, K.}, \bibinfo{author}{Rastegari, M.}, \bibinfo{author}{Farhadi, A.} \& \bibinfo{author}{Mottaghi, R.}
\newblock \bibinfo{title}{Ok-vqa: A visual question answering benchmark requiring external knowledge}.
\newblock In \emph{\bibinfo{booktitle}{Proceedings of the IEEE/cvf conference on computer vision and pattern recognition}}, \bibinfo{pages}{3195--3204} (\bibinfo{year}{2019}).

\bibitem{wang2015explicit}
\bibinfo{author}{Wang, P.}, \bibinfo{author}{Wu, Q.}, \bibinfo{author}{Shen, C.}, \bibinfo{author}{Hengel, A. v.~d.} \& \bibinfo{author}{Dick, A.}
\newblock \bibinfo{title}{Explicit knowledge-based reasoning for visual question answering}.
\newblock In \emph{\bibinfo{booktitle}{Proceedings of the Twenty-Sixth International Joint Conference on Artificial Intelligence}}, \bibinfo{pages}{1290--1296} (\bibinfo{year}{2017}).

\bibitem{wang2017fvqa}
\bibinfo{author}{Wang, P.}, \bibinfo{author}{Wu, Q.}, \bibinfo{author}{Shen, C.}, \bibinfo{author}{Dick, A.} \& \bibinfo{author}{Van Den~Hengel, A.}
\newblock \bibinfo{journal}{\bibinfo{title}{Fvqa: Fact-based visual question answering}}.
\newblock {\emph{\JournalTitle{IEEE transactions on pattern analysis and machine intelligence}}} \textbf{\bibinfo{volume}{40}}, \bibinfo{pages}{2413--2427} (\bibinfo{year}{2017}).

\bibitem{chen2021zero}
\bibinfo{author}{Chen, Z.} \emph{et~al.}
\newblock \bibinfo{title}{Zero-shot visual question answering using knowledge graph}.
\newblock In \emph{\bibinfo{booktitle}{The Semantic Web--ISWC 2021: 20th International Semantic Web Conference, ISWC 2021, Virtual Event, October 24--28, 2021, Proceedings 20}}, \bibinfo{pages}{146--162} (\bibinfo{organization}{Springer}, \bibinfo{year}{2021}).

\bibitem{wu2023resolving}
\bibinfo{author}{Wu, S.}, \bibinfo{author}{Zhao, G.} \& \bibinfo{author}{Qian, X.}
\newblock \bibinfo{journal}{\bibinfo{title}{Resolving zero-shot and fact-based visual question answering via enhanced fact retrieval}}.
\newblock {\emph{\JournalTitle{IEEE Transactions on Multimedia}}}  (\bibinfo{year}{2023}).

\bibitem{brown2020language}
\bibinfo{author}{Brown, T.} \emph{et~al.}
\newblock \bibinfo{journal}{\bibinfo{title}{Language models are few-shot learners}}.
\newblock {\emph{\JournalTitle{Advances in neural information processing systems}}} \textbf{\bibinfo{volume}{33}}, \bibinfo{pages}{1877--1901} (\bibinfo{year}{2020}).

\bibitem{liang2024toa}
\bibinfo{author}{Liang, M.}, \bibinfo{author}{Wu, Y.} \emph{et~al.}
\newblock \bibinfo{journal}{\bibinfo{title}{Toa: Task-oriented active vqa}}.
\newblock {\emph{\JournalTitle{Advances in Neural Information Processing Systems}}} \textbf{\bibinfo{volume}{36}} (\bibinfo{year}{2024}).

\bibitem{kang2024chatmof}
\bibinfo{author}{Yeonghun, K.} \& \bibinfo{author}{Jihan, K.}
\newblock \bibinfo{journal}{\bibinfo{title}{Chatmof: an artificial intelligence system for predicting and generating metal-organic frameworks using large language models}}.
\newblock {\emph{\JournalTitle{Nature Communications and others}}}  (\bibinfo{year}{2024}).

\bibitem{cai2024using}
\bibinfo{author}{Shanqing, C.} \emph{et~al.}
\newblock \bibinfo{journal}{\bibinfo{title}{Using large language models to accelerate communication for eye gaze typing users with als}}.
\newblock {\emph{\JournalTitle{Nature Communications and others}}}  (\bibinfo{year}{2024}).

\bibitem{wei2022chain}
\bibinfo{author}{Wei, J.} \emph{et~al.}
\newblock \bibinfo{journal}{\bibinfo{title}{Chain-of-thought prompting elicits reasoning in large language models}}.
\newblock {\emph{\JournalTitle{Advances in neural information processing systems}}} \textbf{\bibinfo{volume}{35}}, \bibinfo{pages}{24824--24837} (\bibinfo{year}{2022}).

\bibitem{lan2023improving}
\bibinfo{author}{Lan, Y.} \emph{et~al.}
\newblock \bibinfo{title}{Improving zero-shot visual question answering via large language models with reasoning question prompts}.
\newblock In \emph{\bibinfo{booktitle}{Proceedings of the 31st ACM International Conference on Multimedia}}, \bibinfo{pages}{4389--4400} (\bibinfo{year}{2023}).

\bibitem{anderson2018bottom}
\bibinfo{author}{Anderson, P.} \emph{et~al.}
\newblock \bibinfo{title}{Bottom-up and top-down attention for image captioning and visual question answering}.
\newblock In \emph{\bibinfo{booktitle}{Proceedings of the IEEE conference on computer vision and pattern recognition}}, \bibinfo{pages}{6077--6086} (\bibinfo{year}{2018}).

\bibitem{kim2018bilinear}
\bibinfo{author}{Kim, J.-H.}, \bibinfo{author}{Jun, J.} \& \bibinfo{author}{Zhang, B.-T.}
\newblock \bibinfo{journal}{\bibinfo{title}{Bilinear attention networks}}.
\newblock {\emph{\JournalTitle{Advances in neural information processing systems}}} \textbf{\bibinfo{volume}{31}} (\bibinfo{year}{2018}).

\bibitem{yang2016stacked}
\bibinfo{author}{Yang, Z.}, \bibinfo{author}{He, X.}, \bibinfo{author}{Gao, J.}, \bibinfo{author}{Deng, L.} \& \bibinfo{author}{Smola, A.}
\newblock \bibinfo{title}{Stacked attention networks for image question answering}.
\newblock In \emph{\bibinfo{booktitle}{Proceedings of the IEEE conference on computer vision and pattern recognition}}, \bibinfo{pages}{21--29} (\bibinfo{year}{2016}).

\bibitem{ramesh2021zero}
\bibinfo{author}{Ramesh, A.} \emph{et~al.}
\newblock \bibinfo{title}{Zero-shot text-to-image generation}.
\newblock In \emph{\bibinfo{booktitle}{International conference on machine learning}}, \bibinfo{pages}{8821--8831} (\bibinfo{organization}{Pmlr}, \bibinfo{year}{2021}).

\bibitem{lu2016hierarchical}
\bibinfo{author}{Lu, J.}, \bibinfo{author}{Yang, J.}, \bibinfo{author}{Batra, D.} \& \bibinfo{author}{Parikh, D.}
\newblock \bibinfo{journal}{\bibinfo{title}{Hierarchical question-image co-attention for visual question answering}}.
\newblock {\emph{\JournalTitle{Advances in neural information processing systems}}} \textbf{\bibinfo{volume}{29}} (\bibinfo{year}{2016}).

\bibitem{fei2022brivl}
\bibinfo{author}{Nanyi, F.} \emph{et~al.}
\newblock \bibinfo{journal}{\bibinfo{title}{Towards artificial general intelligence via a multimodal foundation model}}.
\newblock {\emph{\JournalTitle{Nature Communications and others}}}  (\bibinfo{year}{2022}).

\bibitem{guo2023images}
\bibinfo{author}{Guo, J.} \emph{et~al.}
\newblock \bibinfo{title}{From images to textual prompts: Zero-shot visual question answering with frozen large language models}.
\newblock In \emph{\bibinfo{booktitle}{Proceedings of the IEEE/CVF Conference on Computer Vision and Pattern Recognition}}, \bibinfo{pages}{10867--10877} (\bibinfo{year}{2023}).

\bibitem{gupta2023visual}
\bibinfo{author}{Gupta, T.} \& \bibinfo{author}{Kembhavi, A.}
\newblock \bibinfo{title}{Visual programming: Compositional visual reasoning without training}.
\newblock In \emph{\bibinfo{booktitle}{Proceedings of the IEEE/CVF Conference on Computer Vision and Pattern Recognition}}, \bibinfo{pages}{14953--14962} (\bibinfo{year}{2023}).

\bibitem{suris2023vipergpt}
\bibinfo{author}{Sur{\'\i}s, D.}, \bibinfo{author}{Menon, S.} \& \bibinfo{author}{Vondrick, C.}
\newblock \bibinfo{title}{Vipergpt: Visual inference via python execution for reasoning}.
\newblock In \emph{\bibinfo{booktitle}{Proceedings of the IEEE/CVF International Conference on Computer Vision}}, \bibinfo{pages}{11888--11898} (\bibinfo{year}{2023}).

\bibitem{khot2022decomposed}
\bibinfo{author}{Khot, T.} \emph{et~al.}
\newblock \bibinfo{title}{Decomposed prompting: A modular approach for solving complex tasks}.
\newblock In \emph{\bibinfo{booktitle}{Decomposed prompting: A modular approach for solving complex tasks}} (\bibinfo{year}{2023}).

\bibitem{huang2022inner}
\bibinfo{author}{Huang, W.} \emph{et~al.}
\newblock \bibinfo{title}{Inner monologue: Embodied reasoning through planning with language models}.
\newblock In \emph{\bibinfo{booktitle}{Proceedings of Machine Learning Research}}, \bibinfo{pages}{1769--1782} (\bibinfo{year}{2022}).

\bibitem{zhang2021vinvl}
\bibinfo{author}{Zhang, P.} \emph{et~al.}
\newblock \bibinfo{title}{Vinvl: Revisiting visual representations in vision-language models}.
\newblock In \emph{\bibinfo{booktitle}{Proceedings of the IEEE/CVF conference on computer vision and pattern recognition}}, \bibinfo{pages}{5579--5588} (\bibinfo{year}{2021}).

\bibitem{he2020momentum}
\bibinfo{author}{He, K.}, \bibinfo{author}{Fan, H.}, \bibinfo{author}{Wu, Y.}, \bibinfo{author}{Xie, S.} \& \bibinfo{author}{Girshick, R.}
\newblock \bibinfo{title}{Momentum contrast for unsupervised visual representation learning}.
\newblock In \emph{\bibinfo{booktitle}{Proceedings of the IEEE/CVF conference on computer vision and pattern recognition}}, \bibinfo{pages}{9729--9738} (\bibinfo{year}{2020}).

\bibitem{tan2019lxmert}
\bibinfo{author}{Tan, H.} \& \bibinfo{author}{Bansal, M.}
\newblock \bibinfo{title}{Lxmert: Learning cross-modality encoder representations from transformers}.
\newblock In \emph{\bibinfo{booktitle}{Empirical Methods in Natural Language Processing}}, \bibinfo{pages}{5100--5111} (\bibinfo{year}{2019}).

\bibitem{he2016deep}
\bibinfo{author}{He, K.}, \bibinfo{author}{Zhang, X.}, \bibinfo{author}{Ren, S.} \& \bibinfo{author}{Sun, J.}
\newblock \bibinfo{title}{Deep residual learning for image recognition}.
\newblock In \emph{\bibinfo{booktitle}{Proceedings of the IEEE conference on computer vision and pattern recognition}}, \bibinfo{pages}{770--778} (\bibinfo{year}{2016}).

\bibitem{deng2009imagenet}
\bibinfo{author}{Deng, J.} \emph{et~al.}
\newblock \bibinfo{title}{Imagenet: A large-scale hierarchical image database}.
\newblock In \emph{\bibinfo{booktitle}{2009 IEEE conference on computer vision and pattern recognition}}, \bibinfo{pages}{248--255} (\bibinfo{organization}{Ieee}, \bibinfo{year}{2009}).

\bibitem{ren2015faster}
\bibinfo{author}{Ren, S.}, \bibinfo{author}{He, K.}, \bibinfo{author}{Girshick, R.} \& \bibinfo{author}{Sun, J.}
\newblock \bibinfo{journal}{\bibinfo{title}{Faster r-cnn: Towards real-time object detection with region proposal networks}}.
\newblock {\emph{\JournalTitle{Advances in neural information processing systems}}} \textbf{\bibinfo{volume}{28}} (\bibinfo{year}{2015}).

\bibitem{lin2014microsoft}
\bibinfo{author}{Lin, T.-Y.} \emph{et~al.}
\newblock \bibinfo{title}{Microsoft coco: Common objects in context}.
\newblock In \emph{\bibinfo{booktitle}{Computer Vision--ECCV 2014: 13th European Conference, Zurich, Switzerland, September 6-12, 2014, Proceedings, Part V 13}}, \bibinfo{pages}{740--755} (\bibinfo{organization}{Springer}, \bibinfo{year}{2014}).

\bibitem{huang2015bidirectional}
\bibinfo{author}{Huang, Z.}, \bibinfo{author}{Xu, W.} \& \bibinfo{author}{Yu, K.}
\newblock \bibinfo{journal}{\bibinfo{title}{Bidirectional lstm-crf models for sequence tagging}}.
\newblock {\emph{\JournalTitle{arXiv preprint arXiv:1508.01991}}}  (\bibinfo{year}{2015}).

\bibitem{zhang2023opt}
\bibinfo{author}{Zhang, S.} \emph{et~al.}
\newblock \bibinfo{journal}{\bibinfo{title}{Opt: Open pre-trained transformer language models, 2022}}.
\newblock {\emph{\JournalTitle{URL https://arxiv. org/abs/2205.01068}}} \textbf{\bibinfo{volume}{3}}, \bibinfo{pages}{19--0} (\bibinfo{year}{2023}).

\end{thebibliography}

\section*{Competing interests}
The authors declare no competing interests.

% \section*{Additional information}
% \textbf{Supplementary information} The online version contains supplementary material available at .\\
% ~\\
% \textbf{Correspondence} and requests for materials should be addressed to Qian Tao.\\
% ~\\
% \textbf{Peer review information} Nature Communications thanks xxx, and the other, anonymous, reviewer(s) for their contribution to the peer review of this work. A peer review file is available.\\
% ~\\
% \textbf{Reprints and permissions information} is available at
% \url{http://www.nature.com/reprints}.\\
% ~\\
% \textbf{Publisher’s note} Springer Nature remains neutral with regard to jurisdictional claims in published maps and institutional affiliations.\\
% ~\\
% \textbf{Open Access} This article is licensed under a Creative Commons Attribution 4.0 International License, which permits use, sharing,
% adaptation, distribution and reproduction in any medium or format, as
% long as you give appropriate credit to the original author(s) and the
% source, provide a link to the Creative Commons license, and indicate if
% changes were made. The images or other third party material in this
% article are included in the article’s Creative Commons license, unless
% indicated otherwise in a credit line to the material. If material is not
% included in the article’s Creative Commons license and your intended
% use is not permitted by statutory regulation or exceeds the permitted
% use, you will need to obtain permission directly from the copyright
% holder. To view a copy of this license, visit \url{http://creativecommons.org/licenses/by/4.0/}. \\

\end{document}